# How Predictable is Your State? Leveraging Lexical and Contextual Information for Predicting Legislative Floor Action at the State Level

**Vlad Eidelman**    **Anastassia Kornilova**    **Daniel Argyle**

FiscalNote Inc.
Washington DC
{vlad,anastassia,daniel}@fiscalnote.com

## Abstract

Modeling U.S. Congressional legislation and roll-call votes has received significant attention in previous literature. However, while legislators across 50 state governments and D.C. propose over 100,000 bills each year, and on average enact over 30% of them, state level analysis has received relatively less attention due in part to the difficulty in obtaining the necessary data. Since each state legislature is guided by their own procedures, politics and issues, however, it is difficult to qualitatively asses the factors that affect the likelihood of a legislative initiative succeeding. Herein, we present several methods for modeling the likelihood of a bill receiving floor action across all 50 states and D.C. We utilize the lexical content of over 1 million bills, along with contextual legislature and legislator derived features to build our predictive models, allowing a comparison of the factors that are important to the lawmaking process. Furthermore, we show that these signals hold complementary predictive power, together achieving an average improvement in accuracy of 18% over state specific baselines.

## 1 Introduction

Federal institutions in the U.S., like Congress and the Supreme Court, play a significant role in lawmaking, and in many observable ways define our legal system. Thus, as data and computational resources have become more readily available, political scientists have increasingly been adopting quantitative methods focused on understanding these entities and the role they play in our society (Katz et al., 2017; Poole and Rosenthal, 2007; Slapin and Proksch, 2008; Lauderdale and Clark, 2014).

Although many issues are legislated and regulated primarily at the federal level, state governments have significant power over certain areas. An increasing number of important issues are being decided at the state or local levels, especially in emerging industries and technologies, such as the gig economy and autonomous vehicles (Hedge, 1998). Moreover, there are 535 members of Congress who introduce over 10,000 pieces of legislation a session,[1] of which less than 5% is enacted. Similar dynamics exist at the state level, except on a much broader scale. There are over 7,000 state legislators, in aggregate introducing over 100,000 pieces of legislation, with over 30% being enacted. In order to be enacted, every bill must pass through one or more legislative committees and be considered on the chamber floor, a process we refer to as receiving floor action. This process is one of the most pivotal steps during lawmaking (Rosenthal, 1974; Hamm, 1980; Francis, 1989; Rakoff and Sarner, 1975), as on average, only 41% of bills receive floor action, with most legislation languishing in the committees.

Legislative policymaking decisions are extremely complex, and are influenced by a myriad of factors, ranging from the content of the legislation, to legislators' personal characteristics, such as profession, religion, and party and ideological affiliations, to their constituents' demographics, to governor agendas, to interest group activities, and to world events (Canfield-Davis et al., 2010; Hicks and Smith, 2009; Talbert and Potoski, 2002).

Despite this complexity, in this paper we present an approach to better understand state lawmaking dynamics and the legislative process by focusing on the task of predicting the likelihood that legislation will reach the floor in each state. As there are many dimensions underlying the content of the legislation, such as the policy area and ideology of the sponsor (Linder et al., 2018), that may affect the likelihood of floor action, in addition to text we examine several established contextual legislature and legislator derived features. To the best of our knowledge, this is the first work quantitatively modeling the floor action process across all 50 states and using the text of legislation alongside traditional contextual information.

---

[1] A session is the period of time a legislative body is actively enacting legislation, usually one to two years.

## 2 Related Work

Much of the work analyzing the federal legislature is aimed at understanding legislator preferences through the use of voting patterns. One of the most popular techniques in political science is the application of spatial, or ideal point, models built from voting records (Poole and Rosenthal, 1985; Poole and Rosenthal, 2007), that is often used to represent unidimensional or multidimensional ideological stances (Clinton et al., 2004). However, there is also an increasing literature examining broader legislative dynamics, such as measuring legislative effectiveness (Harbridge, 2016), evaluating the impact of legislation on stock prices using legislators' constituents (Cohen et al., 2012), creating cosponsorship networks (Fowler, 2006), and examining the role of lobbying (Bertrand et al., 2018; Matthew et al., 2013),

In recent years a variety of primary and secondary textual legal data, such as legislation, floor debates, and committee transcripts, has become increasingly available, enabling the NLP community to create richer multidimensional ideal point estimation (Gerrish and Blei, 2011; Nguyen et al., 2015; Kornilova et al., 2018), and examine ideology detection from political speech (Iyyer et al., 2014), voting prediction from debates (Thomas et al., 2006), committee referral (Yano et al., 2012), and enactment (Nay, 2016).

While there is also an increasing amount of state legislative research, states have received significantly less attention (Hamm et al., 2014). One major reason for this is that quantitative methods require data, and the availability of data for Congress far exceeds that of states. In fact, Yano et al. (2012) noted "When we consider a larger goal of understanding legislative behavior across many legislative bodies (e.g., states in the U.S., other nations, or international bodies), the challenge of creating and maintaining such reliable, clean, and complete databases seems insurmountable." Thus, while there has been scholarship quantifying the role of committees, it has been limited in scope, to a few sessions or states, or reliant on survey data (Francis, 1989; Rakoff and Sarner, 1975; Rosenthal, 1974; Hamm, 1980). More recently, as different kinds of state data has become more accessible, it has enabled studying the affect of interest groups on legislative activity (Gray and Lowery, 1995), the application of spatial models (Shor et al., 2010; Shor and McCarty, 2011), and comparisons of textual similarity (Linder et al., 2018).

The contribution of this work is to continue building a broader understanding of state legislative dynamics by evaluating how predictable state lawmaking is, and what factors influence that process. We create a novel task, predicting the likelihood of legislation receiving floor action, and utilize a corpus of over 1 million bills to build computational models of all 50 states and D.C. We present several baseline models utilizing various features, and show that combining the legislative and legislator contextual information with the text content of bills consistently provides the best predictions. Our analysis considers various factors and their respective importance in the predictive models across the states, showing that although there are some consistent patterns, there are many variations and differences in what affects the likelihood in each state.

## 3 Data

There is state-to-state variation in the legislative procedure of how a bill becomes law, but the path is largely similar. Legislation is introduced by one or more members of the legislature in their respective chamber,[2] and assigned to one or more standing subject committees.[3] Committees are made up of a subset of members of their respective chamber, and are chaired by the majority party. Once in committee,

---

[2] All legislatures are bicameral, with either a House or Assembly as the lower chamber, and the Senate as the upper chamber, except D.C. and Nebraska, which are unicameral.

[3] Depending on the state, other groups can introduce legislation, including legislative committees, legislative delegations, the governor, or non-elected individuals. For the purpose of this work we focus on legislator sponsored legislation.

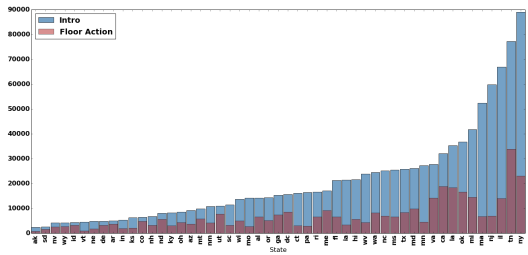 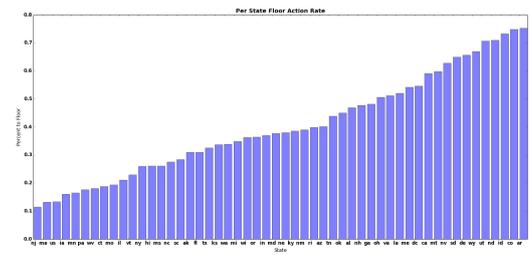

(a) Number of bills introduced and receiving floor action for each state.

(b) Percent of bills reaching floor per state.

Figure 1: Dataset characteristics.

legislation is subject to debate and amendment only by the committee members, with the successful outcome being a favorable referral, or a recommendation, to be considered by the full chamber on the floor.

The primary data we use to model floor action was scraped directly from each state legislatures' website. For each state, we downloaded legislation, committee, and legislator pages for all sessions that were publicly accessible. Legislation pages were automatically parsed to determine legislative contextual metadata, which includes bill text versions, sponsors, committee assignments, and the timeline of actions. Legislator pages were parsed to obtain sponsor contextual metadata, which includes party affiliation, committee assignments, and committee roles.

As states demarcate legislative status in the timeline of actions differently, we automatically map and normalize all textual descriptions of legislative actions to a finite set of statuses.[4] These statuses are used to determine whether a piece of legislation survived committee and received a floor action, or consideration on the floor. All bills having a status of passed in their introductory chamber, or having had a recorded floor vote are treated as positive examples, while any status prior to passed is considered failed, including legislation that was reported out of committee but not considered on the floor.

Finally, since each state follows their own conventions with regard to classifying the type of legislation, we normalize all legislation across states to two types: resolutions and bills.[5]

Figure 1a shows the total number of bills introduced and receiving floor action for each state. In total, our dataset consists of 1.3 million pieces of state legislation, broken into 1 million bills, with 360k receiving floor action, at an average rate across states of 41%, and 275k resolutions, with 210k receiving floor action. On average, we have 10 legislative sessions of data per state.[6] As bills represent substantive legislation, with a much lower floor action rate, while resolutions are much more likely to receive floor action, for the rest of this paper we focus on bills only, and refer to bills and legislation interchangeably. We include 15 sessions of U.S federal legislation in our data for comparative purposes, with 23k of 172k bills receiving floor action.

Figure 1b presents the percent of bills receiving floor action. It is interesting to note the difference in difficulty for legislation to receive floor action in different states. For example, in New Jersey and Massachusetts, fewer than 15% of bills reach the floor, whereas 75% do in Colorado and Arkansas.[7]

## 4 Methods

### 4.1 Models

In order to not only be able to predict, but also examine the importance of features to our prediction, we chose three relatively interpretable models for our modeling framework. Formally, let our training data $(\boldsymbol{X}, \boldsymbol{Y})$ consist of $n$ pairs $(\boldsymbol{x}_i, y_i)_{i=1}^n$ where, each $\boldsymbol{x}_i$ is a bill and $y_i$ a binary indicator of whether $\boldsymbol{x}_i$ received floor action. Let $\mathbf{f}(\boldsymbol{x}_i)$ be a feature vector representation of $\boldsymbol{x}_i$, and $\mathbf{w}$ the parameter vector indicating the weight of each feature learned by the model.

---

[4]The normalized statuses include introduced, assigned to committee, reported from committee, and passed.

[5]Resolutions are pieces of legislation of type appointment, resolution, joint resolution, concurrent resolution, joint memorial, memorial, proclamation, nomination. Bills are those of type bill, amendment, urgency, appropriation, tax levy, or constitutional amendment.

[6]Full data statistics are given in Table 7 in Appendix A.

[7]Our average across states, chambers, and sessions is in line with previous single state and session findings; in examining five states Rosenthal (1974) found between 34% and 73% of legislation did not survive committee.

The first two models are linear classifiers, where the prediction of floor action, $\hat{y}_i$, is given by `sign`$(\mathbf{w}^\top \mathbf{f}(\boldsymbol{x}))$. The first is a regularized conditional log-linear model $p_\mathbf{w}(y \mid \boldsymbol{x})$:

$$p_\mathbf{w}(y \mid \boldsymbol{x}) = \frac{\exp\{\mathbf{w}^\top \mathbf{f}(\boldsymbol{x})\}}{Z(\boldsymbol{x})} \tag{1}$$

where $Z(\boldsymbol{x})$ is the partition function given by $\sum_y \exp\{\mathbf{w}^\top \mathbf{f}(\boldsymbol{x})\}$. The model optimizes $\mathbf{w}$ according to

$$\min_\mathbf{w} \sum_i^n -\log p_\mathbf{w}(y_i \mid \boldsymbol{x}_i) + \lambda ||\mathbf{w}|| \tag{2}$$

The second model is NBSVM (Wang and Manning, 2012), an interpolation between multinomial Naive Bayes and a support vector machine, which optimizes $\mathbf{w}$ according to:

$$\min_\mathbf{w} C \sum_i^n \max(0, 1 - y_i(\mathbf{w}^\top (\mathbf{f}(\boldsymbol{x}_i) \circ \mathbf{r})))^2 + ||\mathbf{w}||^2 \tag{3}$$

where $\mathbf{r}$ is the log-count ratio of features occurring in positive and negative examples. The third model is non-linear, in the form of a tree-based gradient boosted machine (Friedman, 2000), which optimizes $\mathbf{w}$ according to:

$$\min_\mathbf{w} \sum_i^n l(y_i, \hat{y}_i) + \sum_{k=1}^K \Omega(\mathbf{t}_k) \tag{4}$$

where $K$ is the number of trees, $l$ is the loss function, typically binomial deviance, and $\hat{y}_i$ is given by $\sum_{k=1}^K \mathbf{t}_k(\boldsymbol{x}_i)$ where $\mathbf{t}_k$ is a tree.

We use the `scikit-learn` (Pedregosa et al., 2011) implementation for the log-linear and gradient boosted models, and implemented NBSVM based on the interpolated version in Wang and Manning (2012).

As hyperparameters, such as learning rate and regularization, have a significant impact on model performance, we use Bayesian hyperparameter optimization (Bergstra et al., 2011) to select the optimal hyperparameters for each model on a held-out development set. We used the tree-structured Parzen Estimator (TPE) algorithm implemented in `hyperopt` for our sequential model-based optimization (Bergstra et al., 2013). After individually optimizing hyperparameters and training each of the three base models, we use their outputs to train a meta-ensemble model, a regularized conditional log-linear model, forming a linear combination over their predictions (Breiman, 1996).

As the lawmaking process in each state, and even within each chamber, is different, we divide the problem space by state and chamber, building separate models for each subset. Specifically, we consider each of these as separate problems: upper chamber bills and lower chamber bills. Thus, we have 2 predictions per state, and each prediction is comprised of 4 model outputs, three from the base models, and one from the meta-ensemble, resulting in 400 models.[8]

### 4.2 Features

As there are many dimensions underlying bills that may affect the likelihood of floor action, we compute and utilize several established contextual legislature and legislator derived features. Previous literature has proposed various factors that may affect legislation, including the content of bills,[9] number of and identity of sponsors, extra-legislative forms of support, timing of introduction, leadership's position, seniority, identity of chairperson of the committee, identity of one's own party, and membership of the dominant faction (Hamm, 1980; Rakoff and Sarner, 1975; Harbridge, 2016; Yano et al., 2012).

---
[8] There are only upper chamber bills in D.C. and Nebraska, resulting in 49 states x 2 prediction types + 2 states x 1 prediction type) x 4 models = 400.

[9] In most previous literature the content is determined via a manual analysis of each bill to establish the scope of impact, the complexity, or the incremental nature.

| Feature Type | Description |
|---|---|
| Sponsor | primary and cosponsor(s) identity, primary and cosponsors(s) party affiliation, number of primary and sponsors, number of Republicans, number of Democrats, sponsors bicameral, sponsors bipartisan, sponsor in majority/minority, majority party Republican or Democrat |
| Committee | identity of assigned committee(s), number of committee assignments, number of sponsors members of the committee, sponsor same party as committee chairman, sponsor role on the committee, referral rate of committee(s) |
| Bill | chamber, bill type, session, introductory date, companion bill(s) existence, companion(s) current status. |

Table 1: Contextual feature types and descriptions.

In order to quantitatively evaluate these factors and establish a strong baseline from which to measure the affect of text, we include the contextual features shown in Table 1. These indicator features derived from the sponsors, committees, and bills are meant to capture many of the major factors that are proposed in the literature.[10] To strengthen the representation of legislators in our model beyond the basic features described above, we compute several measures of legislator effectiveness. The effectiveness score is calculated from the sponsoring and cosponsoring activity of each legislator, and meant to represent where they stand in relation to other legislators in successfully passing legislation.[11]

Similar to Harbridge (2016), the score we compute for each legislator is a combination of several partial scores, computed for each important stage of the legislative process. Each legislator gets a score for how many bills they sponsored, getting those bills out of committee, getting them to the floor, passing their own chamber, passing the legislature, and getting enacted. The score for each stage is further broken down by how many of those pieces of legislation were substantive, i.e. bills, attempting a meaningful legal change, versus non-substantive, i.e. resolutions. This results in 12 factors for each individual. To compute a score for each legislator's relative performance at each stage to the other members in the chamber, we create a weighted combination of that legislator's bills and resolutions, where bills get more weight, and compute the ratio based on the weighted contribution of the other members in the chamber. All the stage scores are then combined into a second weighted combination, where each successive stage in the process gets more weight, to get the final score. Finally, the scores are normalized to 0-10. In addition to using the effectiveness scores directly as features, we further compute and discretize several statistics derived from them, including ranks, percentiles, and deviations from the mean thereof.

To further enrich the bill representation beyond contextual information, we utilize the textual content of the bills. The legislation in our collection is comprised of long documents, with an average of 11 thousand words, often containing significant amounts of procedural language and pieces of extant statutes. As this can create additional challenges in identifying the salient points, for this work we chose to focus on a condensed amount of text, specifically the state provided title and description, that average 17 and 18 words, respectively.[12] Both are preprocessed by lowercasing and stemming. We compute the tf-idf weighting for n-grams of size (1,3) on the training data for each prediction task, and select the top 10k

---

[10]Each count based feature, such as number of sponsors, also spawns a number of discretized features, including ranks, percentiles, and deviations from the mean thereof. We automatically compute companion bills using a cosine-based lexical similarity.

[11]This is not a holistic representation of being an effective legislator, as someone may consider themselves effective by not passing anything, or preventing others from doing so. Members may also be highly influential and their support is needed behind the scenes but their names do not appear on the legislation. We can only account for recorded activity. Despite the limitations, we argue this is a fair, if incomplete, assessment of how well the legislator advances their agenda.

[12]Although this is a coarse approximation of the bill content, we believe it should capture the substantive aspects of the bill. Full details of the length of documents in each state are given in Table 7 in Appendix A.

| Condition | Feature Set |
|---|---|
| `combined` | sponsor, committee, bill, text |
| `no_txt` | sponsor, committee, bill |
| `no_txt_spon` | committee, bill |
| `just_txt` | text |
| `just_spon` | sponsor |

Table 2: The five feature settings with contextual and lexical features.

n-grams from the title and description separately.

While we would like to study the predictability of reaching the floor upon first introduction, bills often change after introduction and are updated with additional information. Thus, we limit our features to those available at the time of first introduction.

## 5 Results

In order to clarify the impact that each set of features described in Section 4.2 has on predictive performance, we create five different subsets of features described in Table 2, and train models on each one of them separately.

The first condition, `combined` contains all the contextual and text content features. The second condition, `no_txt`, removes text content, allowing us to study the importance of all contextual features, and by comparing `combined` to `no_txt` we can evaluate if text has any complementary information to contextual features. The third condition, `no_txt_spon` further removes sponsor features, essentially allowing us to study the importance of committee information. By comparing `no_txt` to `no_txt_spon` we can evaluate what sponsors contribute. The fourth and fifth conditions use only sponsor and only text features, respectively, to study the importance of each individually.

All models for a given condition are built from the same training data and feature space. We measure and report several performance metrics of our models using 10-fold cross validation. The baseline model represents guessing the majority class; for some states this means all fail, for others it is all receive floor action, based on the state specific rate.

Although accuracy is informative with respect to how many correct binary decisions the model made, as noted in Bradley (1997) for imbalanced problems such as this, where one class dominates, the baseline accuracy can be very high. As a supplement, it is useful to measure a probabilistic loss, where there is a cost associated with how correct the decision was. Thus, we move beyond pure predictive performance and consider the actual probability distributions created by our models under different conditions. The log-linear and gradient boosted models are probabilistic, while NBSVM is not, thus we train a probability transformation on top of NBSVM using Platts Scaling to obtain probability estimates.

In additional to accuracy, we measure model performance on log-loss and AUROC (area under the receiver operating characteristic curve) (Bradley, 1997). Log-loss, $LL$ is defined as:

$$LL = -\frac{1}{n}\sum_{i=1}^{n} \mathbb{1}(y_i = \hat{y}_i)\log(p_i) + (1 - \mathbb{1}(y_i = \hat{y}_i))\log(1 - p_i) \qquad (5)$$

where $\mathbb{1}(y_i = \hat{y}_i)$ is an binary indicator function equaling 1 if the model prediction $\hat{y}_i$ was correct, and 0 otherwise. $LL$ equals zero for a perfect classifier, and increases with worse probability estimates. Specifically, $LL$ penalizes models more the more confident they are in an incorrect classification.

AUROC allows us to measure the relationship between a model's true positive (TP), how many floor action bills were correctly predicted as floor action, and false positive rate (FP), how many failed bills were predicted as floor action. It is defined by:

$$AUROC = \sum_{i=1}^{N} p(TP)\Delta p(FP) + \frac{1}{2}(\Delta p(TP)\Delta p(FP)) \qquad (6)$$

| Feature Set | Accuracy | | Log-Loss | | AUROC | |
|---|---|---|---|---|---|---|
| | Average | Std Dev | Average | Std Dev | Average | Std Dev |
| `baseline` | 0.68 | 0.1 | 0.6 | 0.09 | 0.5 | 0 |
| `just_txt` | 0.732 | 0.09 | 0.53 | 0.14 | 0.7 | 0.14 |
| `just_spon` | 0.759 | 0.102 | 0.48 | 0.16 | 0.74 | 0.15 |
| `no_txt_spon` | 0.81 | 0.113 | 0.39 | 0.18 | 0.8 | 0.18 |
| `no_txt` | 0.846 | 0.098 | 0.32 | 0.18 | 0.82 | 0.21 |
| `combined` | **0.859** | 0.093 | **0.31** | 0.17 | **0.85** | 0.21 |

Table 3: Average and standard deviation across states on accuracy, log-loss, AUROC for bills on each feature set.

By considering the TP and FP at different values, we can construct an ROC curve. The area under that curve, AUROC, can be interpreted as the probability that the model will rank a uniformly selected positive instance (floor action) higher than a uniformly selected negative instance (failure), or in other words, the average rank of a positive example. A random model will have a AUROC of 0.5, and a 45-degree diagonal curve, while a perfect model will have an AUROC of 1, and be vertical, then horizontal.

Table 3 shows the average accuracy, $LL$, and AUROC with standard deviations for each of the five conditions. The `just_txt` model achieves an accuracy of 73%, outperforming the baseline by 5%, and notably, shows that there is a predictive signal even within the limited amount of text available in the title and descriptions.

To examine where text content is most and least predictive on its own, we disentangle the average performance of the `just_txt` model in Figure 2a, showing the per state and chamber pair change from baseline. The states that improve the most over baseline, with 15% improvement or more using only textual features are Oregon, Oklahoma, Tennessee, D.C., South Carolina, Louisiana (lower), Georgia (lower), and Alabama (lower). On the other hand, text is least predictive in Connecticut, Wyoming, Idaho, New Jersey, Utah (upper), New Hampshire (upper), North Dakota (upper), all underperforming the baseline.

The relatively small improvement over baseline of `just_txt` provides insight into the lawmaking process, raising the possibility that other contextual factors, outside the subject matter of the legislation, such as who the sponsors are and what committee the bill is assigned to, are often more important than the subject of the legislation.

The `just_spon` model achieves an average accuracy of 76%, slightly outperforming `just_txt` with an improvement over baseline of 8%. This further indicates that knowing sponsor related information, without reference to the subject of the legislation, is itself highly predictive. In fact, Figure 2b shows that except for New Hampshire (upper), almost all states achieve gains using sponsor only information, with Oklahoma, Texas, and Ohio achieving gains of 30% or more. The committee information in `no_txt_spon`, which includes the sponsor committee positions, is even more predictive than sponsor and text only, and the addition of sponsors in `no_txt` improves performance by 3.5%.

Including text in the `combined` model further improves performance by 1.3% over `no_txt`, and 18% over the majority class baseline, showing the complementary effects of contextual and lexical information, as this model consistently outperforms all others. Figure 2c shows the per state and chamber pair `baseline` and `combined` model performance. The AUROC performance follows a very similar trajectory.

On $LL$, the model performance follows a similar path, with all models showing improvement in probability estimates from the baseline. $LL$ almost doubles from the `combined` model's 0.31 to 0.6 on `baseline`. This reinforces that the `combined` model makes very confident correct predictions. Including text in the `combined` improves performance slightly over `no_txt`, while having just sponsors or just text decreases the $LL$ to around 0.5.

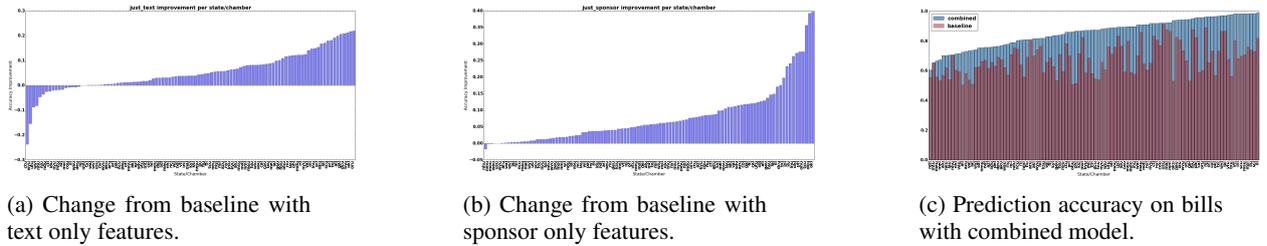

(a) Change from baseline with text only features.

(b) Change from baseline with sponsor only features.

(c) Prediction accuracy on bills with combined model.

Figure 2: Performance improvements for text-only (2a) and sponsor-only models (2b), and combined performance (2c)

## 6 Analysis

All contextual and lexical features considered above are available upon the introduction of a bill, or shortly thereafter,[13] thus the evaluation above indicates how well floor action can be predicted from the day of introduction. However, after the bill is introduced, subsequent legislative actions indicate further contextual information about the legislative process. As it is reasonable to assume these actions carry relevant predictive information, we further examine subsequent events in the legislative process in the `combined+act` feature set. We include a binary feature for the occurrence of amendment introduction and outcomes, votes, committee referral outcomes and readings up to the point of floor action. By comparing `combined+act` to `combined` we can examine how important different events in the legislative procedure are to predicting floor action.

Table 4: Average accuracy, log-loss, AUROC for bills using legislative events post introduction.

| Feature Set | Accuracy | | Log-Loss | | AUROC | |
|---|---|---|---|---|---|---|
| | Average | SD | Average | SD | Average | SD |
| `combined` | 0.859 | 0.093 | 0.31 | 0.17 | 0.85 | 0.21 |
| `combined+act` | **0.94** | 0.059 | **0.16** | 0.12 | **0.97** | 0.04 |

Table 4 shows the results of the `combined+act` feature set. Accuracy improves to 0.94, while $LL$ drops by half to 0.16, confirming that legislative events occurring up to the point point of floor action carry significant complementary information to other contextual factors and are highly indicative of floor action. While `combined+act` confirms the predictive power of procedural factors outside the legislative text, sponsor, and committee assignment, the `combined` model is arguably the most important result, as it indicates how well we can predict on features that are available upon introduction.

Beyond the predictions, we are interested in identifying the different features that contribute to legislative success across the states. As there are both a large number of models, and features in each model, in order to understand the relative predictive importance of contextual and legislature specific dynamics, we choose several previously proposed factors deemed to be important for floor action, and compare the rank and weight they received in each model.

We first examine the median rank and weight given to the following features in the `just_spon` condition across all states: bipartisan, sponsor in minority, sponsor in majority, and the number of sponsors. While many of these contextual features are highly ranked, there are many variations and differences across states. The top half of Table 5 shows the top ten states for which each feature was ranked among the top 20. For example, the bipartisan feature is ranked in the top 5 in Missouri, Virginia, Maine, and Mississippi, accounting for up to 6% of the explanatory power. As a comparison, in South Dakota, Hawaii, Minnesota, Wisconsin, and Pennsylvania, bipartisanship ranked lower than 200. Whether the sponsor is in the minority is important in the U.S. Congress, where it is ranked $6^{th}$, along with Delaware, Tennessee, and West Virginia. Being in the majority accounts for 10% in Kentucky, and 7% in Wisconsin. This aligns with previous literature, as Wisconsin is known to have a strong party system (Hamm, 1980), and indeed we find sponsor in majority and in minority features to be ranked $1^{st}$ and $9^{th}$, respectively, while in Texas, which has a much weaker party system, those features are ranked among the lowest of all states.[14]

---
[13]Some states do not indicate committee assignment immediately, for those we include the first assignment after introduction.

| Feature | Median | Top | Top States | Bottom States |
|---|---|---|---|---|
| Bipartisan | 64 | 11 | mo,va,me,ms,nc,sc,ak,de,wa,us | sd,mn,wi,pa,ut,ne,id,fl,dc,ar |
| in Min | 24 | 20 | de,us,wv,tn,ia,wi,al,nd,md,mi | ca, il,hi,tx,nj,ut,ne,fl,dc,ar |
| in Maj | 23 | 22 | wi,mn,tn,ky,nh,nc,co,il,al,oh | ak,tx,ma,va,ne,nj,me,ut,fl,ar |
| Num Spon | 28 | 20 | co, ut,il,vt,in,ia,or,sd,oh,us | pa, az,wa,wy,nm,nv,va,ms,mn,ar |
| Ranking Mbr | 24 | 15 | ne,vt,ar,ky,us,ga,ok,me,ny,or | ks,mo,mt,oh,pa,ri,tn,ut,va,wy |
| No Cmte Mbr | 17 | 23 | ar,me,ne,il,sd,nc,nv,nm,de,ky | hi,id,ia,ks,mt,oh,pa,tn,ut,wy |
| Members | 6 | 33 | de,ct,me,sd,nc,nv,ok,ny,ga,il | hi,id,ia,ks,mt,oh,pa,tn,ut,wy |

Table 5: Median ranking of across states for bipartisan, sponsor in minority, sponsor in majority, and number of sponsor features for sponsor only model, and having a ranking majority of the committee as a sponsor, not being a committee member as a sponsor, and being a member of the committee as a sponsor in the committee model. The top column indicates how many states have that feature ranked within the top 20 weighted features. The top states lists the ten states where each feature was ranked the highest and was one of the first 20 features. The bottom rows lists the ten states where each feature was ranked the worst.

Similar ranking is presented for committee features in the bottom half of Table 5 in the `no_txt_spon` condition. The committee features play a similarly predictive role, with the sponsors holding membership positions on the committee accounting for over 10% of explanatory power in Delaware, Connecticut, Maine, and South Dakota.

To examine the difference in probability assigned by the models under different conditions, we chose a representative example where neither contextual nor lexical features dominate, as shown in Figure 2, and show the boxplots for Pennsylvania's lower chamber in Figure 3. Each subfigure shows the probability of floor action for legislation that received floor action (pass) and did not (failed). In all cases, the median of the probabilities on legislation that received floor action is higher than the median of the probabilities on legislation that failed. The `combined` models median and mean predictions on bills receiving action are above 90%, and it has the largest difference between the two cases. The `no_txt` model has a similar mean, but the probabilities become more distributed on both pass and fail. Removing sponsors significantly affects the distribution, and shifts the mean lower to 70%. `just_spon` and `just_text` both drop the mean to around 40%.

In addition, we show the calibration curves and distribution of predictions for the same settings in Figure 4 and ROC curves in Figure 5 in Appendix A. All models are well calibrated, closely following the diagonal line. The `combined` model is very confident in its predictions, resembling a bimodal distribution, placing most predictions close to either 0 or 1 probability. `just_spon` has the most distributed probability estimates, while `just_txt` moves the lower part of the distribution slightly forward. The `combined` model is quite accurate, with each subsequent model moving the ROC curve to the right, and thus allowing more false positives to reach the same true positive rate.[15]

Finally, we examine language ranked most and least predictive on the `just_txt` condition for New Mexico, Pennsylvania, and New York in Table 6.[16] Previous literature has proposed several theories on how content affects legislative passage, including that the more redistributive a policy is perceived, the higher in controversy, or the greater in scope, the lower the passage likelihood (Rakoff and Sarner, 1975; Hamm, 1980). While each state has a unique set of issues that are likely to be taken to the floor, and conversely, to be left in committee, there is also evident overlap. In the top phrases, several states contain budgetary issues, expressed with fiscal and appropriation language, as most states have to pass budgetary measures. We also see commendation and procedural language, which is often less contentious. In the bottom phrases, several states have tax related language, and several education related topics.

---

[14] Full rankings and weights are presented in Table 8 in Appendix A.

[15] For comparison to a state with a higher rate of floor action, we include analogous figures for California's lower chamber in Appendix A.

[16] Addition states are presented in Table 9 in Appendix A.

| State | Top Phrases | Bottom Phrases |
| --- | --- | --- |
| New Mexico (upper) | day, campus, recognit, month, defin, alcohol, date, recipi, procur, cours, registr plate, revis, | tax credit chang, enmu, residenti, lobbi, statewid, or, abort, safeti, date for, test for, primari care, analysi, |
| New Mexico (lower) | day, studi, length, citi, of nm, fingerprint, geotherm, fund project, dog, definit, loan for, month, | of game fish, peac, senior citizen, math scienc, transfer of, state fair, self, bachelor, develop tax credit, nmhu, wolf, equip tax, |
| Pennsylvania (upper) | provid for alloc, creation of board, manufactur or, an appropri to, of applic and, medic examin, fiscal offic, for request for, corpor power, within the general, for the offic, | an act amend, as the tax, known the, act provid, known the tax, wage, act prohibit, citizen, of pennsylvania further, tax, youth, requir the depart, |
| Pennsylvania (lower) | or the, contract further, and for special, memori highwai, within the general, in game, first class township, whistleblow, emerg telephon, offens of sexual, for promulg, | act amend titl, an act amend, act provid, known, act prohibit, amend the, pennsylvania, an act provid, code of, act establish, an act relat, the constitut, |
| New York (upper) | fiscal year relat, memori highway, year relat to, implement the health, for retroact real, portion of state, the public protect, implement the public, inc to appli, budget author, program in relat, which are necessari | languag assist, direct the superintend, the develop of, author shall, subsidi, automobil insur, such elect, limit profit, disabl act, polici base, polici to provid |
| New York (lower) | care insur, applic for real, physic educ, fire district elect, establish credit, to file an, abolit or, hous program, the suspens of, are necessari to, the membership of, relat to hous | appropri, fuel and, numer, school ground, vehicular, incom tax for, prohibit public, tag, senat and assembl, on school, on school, class feloni |

Table 6: Top and bottom ranked phrases for New Mexico, Pennsylvania, and New York.

While outside the scope of this work, in future work we hope to explore the differing language identified by the model to help identify important questions about the policymaking process in each state, and allow comparison within states of what successful legislation contains, and across states, of how different issues take shape. In addition, as we only included a limited amount of text, we would like to explore how to incorporate the full body text of legislation effectively.

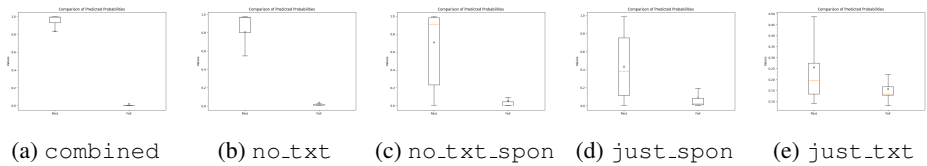

(a) `combined` (b) `no_txt` (c) `no_txt_spon` (d) `just_spon` (e) `just_txt`

Figure 3: Box plot distributions of predicted probabilities for legislation in Pennsylvania lower chamber. The box extends from the lower to upper quartile values of the predictions, with a line at the median, and triangle at the mean.

# 7 Conclusion

In this paper we explored the state legislative process by introducing the task of predicting floor action across all 50 states and D.C. We presented several baseline models and showed that combining contextual information about the legislators and legislatures with bill text consistently provides the best predictions, achieving an accuracy of 86% on which legislation will reach the floor upon first introduction. We further analyzed various factors and their respective importance in the predictive models across the states, gaining a broader understanding of state legislative dynamics. While the factors that influence legislative floor action success are diverse and understandably inconsistent among states, by examining them we can empirically help elucidate the similarities and differences of the policymaking processes.

# A Appendix

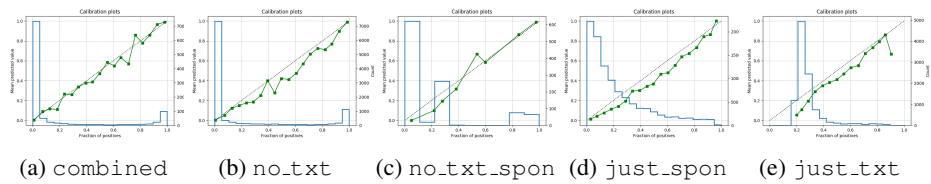

Figure 4: Calibration plots for Pennsylvania lower chamber.

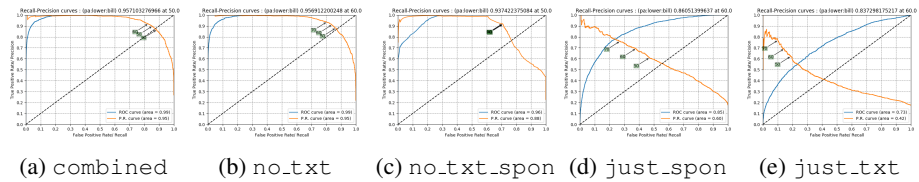

Figure 5: ROC curves and AUC for Pennsylvania lower chamber. Green pointers indicate probability thresholds on the Recall-Precision curve, and the title includes accuracy at the best performing threshold.

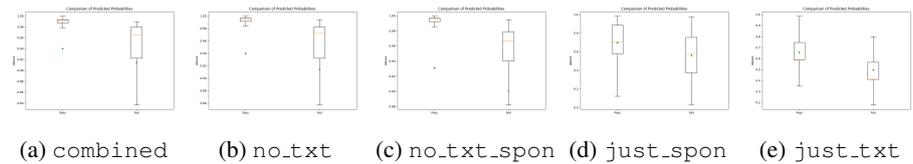

Figure 6: Distributions of predicted probabilities for legislation in California lower chamber that received floor action (pass) and did not (fail) in a box plot. The box extends from the lower to upper quartile values of the predictions, with a line at the median, and triangle at the mean.

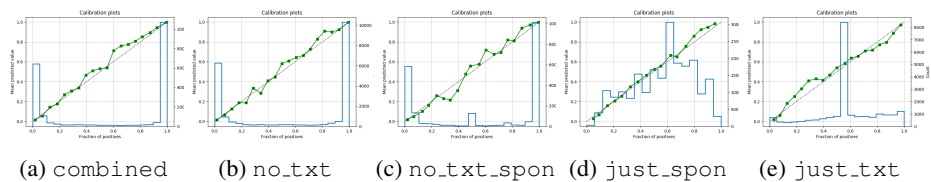

Figure 7: Calibration plots for California lower chamber.

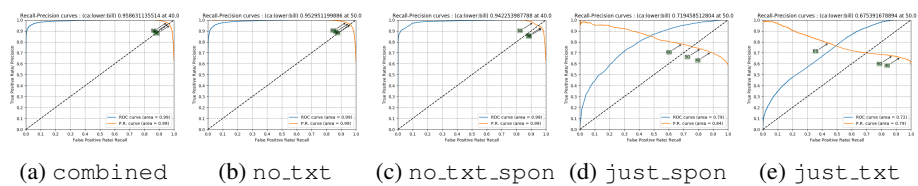

Figure 8: ROC curves and AUC for California lower chamber. Green pointers indicate probability thresholds on the Recall-Precision curve, and the title includes accuracy at the best performing threshold.

| State | Floor Action | Introduced | Rate | Sessions | Title | Desc | Body |
|---|---|---|---|---|---|---|---|
| al | 6697 | 14327 | 0.467 | 16 | 21 | - | 11280 |
| ak | 781 | 2527 | 0.309 | 4 | 34 | - | 12054 |
| az | 3719 | 9308 | 0.4 | 24 | 4 | - | 13284 |
| ar | 3809 | 5076 | 0.75 | 8 | 17 | - | 7297 |
| ca | 18978 | 32143 | 0.59 | 17 | 5 | 186 | 16631 |
| co | 4808 | 6428 | 0.748 | 11 | 5 | 20 | 10671 |
| ct | 3044 | 16236 | 0.187 | 8 | 12 | 19 | 4748 |
| de | 3185 | 4858 | 0.656 | 6 | 19 | 89 | 9327 |
| dc | 8515 | 15593 | 0.546 | 9 | 11 | - | 7083 |
| fl | 6592 | 21298 | 0.31 | 15 | 6 | 29 | 16471 |
| ga | 7416 | 15379 | 0.482 | 13 | 8 | 43 | 5654 |
| hi | 5630 | 21615 | 0.26 | 5 | 9 | 28 | 6910 |
| id | 3259 | 4446 | 0.733 | 8 | 40 | - | 10458 |
| il | 14106 | 66926 | 0.211 | 10 | 3 | 46 | 7022 |
| in | 1958 | 5291 | 0.37 | 4 | 5 | 55 | 16918 |
| ia | 3434 | 21457 | 0.16 | 7 | 28 | - | 12740 |
| ks | 2123 | 6324 | 0.336 | 6 | 12 | - | 22294 |
| ky | 3149 | 8185 | 0.385 | 20 | 51 | 52 | 14218 |
| la | 18346 | 35277 | 0.52 | 32 | 20 | - | 7454 |
| me | 9268 | 17095 | 0.542 | 8 | 14 | - | 5583 |
| md | 9857 | 26125 | 0.377 | 14 | 10 | 59 | 8107 |
| ma | 6862 | 52467 | 0.131 | 7 | 11 | 26 | 14535 |
| mi | 14520 | 41730 | 0.348 | 11 | 27 | - | 10929 |
| mn | 4494 | 27240 | 0.165 | 10 | 11 | - | 10251 |
| ms | 6621 | 25450 | 0.26 | 21 | 12 | 77 | 14475 |
| mo | 2736 | 14143 | 0.193 | 8 | 18 | - | 16486 |
| mt | 5910 | 9905 | 0.597 | 8 | 7 | - | 13758 |
| ne | 1837 | 4829 | 0.38 | 6 | 12 | - | 11775 |
| nv | 2614 | 4163 | 0.628 | 7 | 14 | 12 | 21165 |
| nh | 3243 | 6793 | 0.477 | 6 | 12 | - | 3269 |
| nj | 6900 | 59861 | 0.115 | 8 | 16 | - | 12883 |
| nm | 4253 | 10909 | 0.39 | 8 | 5 | - | 11639 |
| ny | 23071 | 89072 | 0.259 | 4 | 16 | 23 | 8913 |
| nc | 6922 | 25152 | 0.275 | 10 | 4 | - | 7039 |
| nd | 5735 | 8089 | 0.709 | 9 | 42 | 3 | 5847 |
| oh | 4356 | 8605 | 0.506 | 9 | 8 | - | 33216 |
| ok | 16579 | 36827 | 0.45 | 10 | 13 | - | 10333 |
| or | 5240 | 14404 | 0.364 | 12 | 9 | 54 | 14500 |
| pa | 2887 | 16414 | 0.176 | 5 | 32 | - | 5466 |
| ri | 6596 | 16584 | 0.398 | 5 | 31 | - | 8838 |
| sc | 3269 | 11532 | 0.283 | 6 | 77 | 7 | 6255 |
| sd | 1647 | 2539 | 0.649 | 9 | 16 | - | 6437 |
| tn | 33936 | 77331 | 0.439 | 12 | 30 | - | 3256 |
| tx | 8371 | 25771 | 0.325 | 9 | 16 | - | 5145 |
| ut | 7816 | 11072 | 0.706 | 23 | 7 | - | 24623 |
| vt | 1035 | 4520 | 0.229 | 4 | 14 | - | 6046 |
| va | 14215 | 27813 | 0.511 | 24 | 9 | 37 | 8310 |
| wa | 8317 | 24578 | 0.338 | 6 | 10 | - | 14735 |
| wv | 4308 | 23917 | 0.18 | 12 | 12 | 10 | 11501 |
| wi | 4982 | 13761 | 0.362 | 14 | 40 | 2 | 12090 |
| wy | 2825 | 4223 | 0.669 | 11 | 4 | 38 | 7032 |
| us | 22973 | 172921 | 0.133 | 15 | 14 | 178 | 14043 |

Table 7: Data statistics for number of bills introduced and receiving floor action for each state. Word counts are given for title, description, and bill body.

| State | Bipartisan | | in Min | | in Maj | | Num Spon | | AP Eff | | BP Eff | |
|---|---|---|---|---|---|---|---|---|---|---|---|---|
| | Rank | Weight | Rank | Weight | Rank | Weight | Rank | Weight | Rank | Weight | Rank | Weight |
| al | 57 | 0.003 | 12 | 0.012 | 5 | 0.017 | 54 | 0.003 | 17 | 0.015 | 95 | 0.003 |
| ak | 9 | 0.029 | 22 | 0.009 | 57 | 0.002 | 62 | 0.006 | 3 | 0.059 | 3 | 0.046 |
| az | 133 | 0.003 | 46 | 0.004 | 28 | 0.008 | 69 | 0.006 | 89 | 0.003 | 11 | 0.012 |
| ar | - | - | - | - | - | - | - | - | 0 | 0.326 | 1 | 0.236 |
| ca | 43 | 0.006 | 108 | 0.003 | 29 | 0.007 | 13 | 0.01 | 70 | 0.003 | 3 | 0.027 |
| co | 49 | 0.006 | 19 | 0.014 | 3 | 0.025 | 0 | 0.137 | 94 | 0.002 | 135 | 0.0 |
| ct | 128 | 0.001 | 75 | 0.003 | 26 | 0.008 | 58 | 0.004 | 210 | 0.001 | - | - |
| de | 13 | 0.013 | 3 | 0.023 | 36 | 0.013 | 22 | 0.009 | 79 | 0.003 | 115 | 0.002 |
| dc | - | - | - | - | 16 | 0.015 | 33 | 0.011 | - | - | - | - |
| fl | - | - | - | - | - | - | 27 | 0.012 | 1 | 0.072 | 0 | 0.137 |
| ga | 80 | 0.003 | 23 | 0.01 | 9 | 0.015 | 32 | 0.006 | 303 | 0.0 | 291 | 0.0 |
| hi | 224 | 0.0 | 120 | 0.001 | 26 | 0.006 | 12 | 0.009 | 136 | 0.001 | 0 | 0.321 |
| id | - | - | 24 | 0.001 | 28 | 0.001 | 24 | 0.002 | 14 | 0.018 | - | - |
| il | 159 | 0.001 | 109 | 0.015 | 3 | 0.037 | 2 | 0.015 | 179 | 0.0 | 24 | 0.006 |
| in | 43 | 0.005 | 101 | 0.002 | 50 | 0.009 | 2 | 0.044 | 42 | 0.007 | 48 | 0.004 |
| ia | 98 | 0.001 | 8 | 0.026 | 14 | 0.015 | 3 | 0.086 | 192 | 0.0 | - | - |
| ks | 112 | 0.0 | 69 | 0.001 | 32 | 0.007 | 16 | 0.016 | 39 | 0.009 | 61 | 0.005 |
| ky | 35 | 0.004 | 38 | 0.006 | 2 | 0.103 | 67 | 0.004 | - | - | - | - |
| la | 32 | 0.004 | 30 | 0.005 | 24 | 0.006 | 14 | 0.053 | 23 | 0.005 | 35 | 0.003 |
| me | 5 | 0.055 | 21 | 0.012 | - | - | 21 | 0.013 | 40 | 0.001 | - | - |
| md | 36 | 0.006 | 14 | 0.011 | 22 | 0.008 | 33 | 0.006 | 4 | 0.049 | 7 | 0.029 |
| ma | 151 | 0.004 | 69 | 0.003 | 66 | 0.003 | 16 | 0.009 | 35 | 0.009 | 3 | 0.045 |
| mi | 80 | 0.002 | 14 | 0.01 | 10 | 0.033 | 28 | 0.006 | 98 | 0.002 | 138 | 0.001 |
| mn | 231 | 0.0 | 60 | 0.003 | 1 | 0.046 | 283 | 0.0 | 156 | 0.003 | 140 | 0.003 |
| ms | 5 | 0.029 | 30 | 0.01 | 32 | 0.018 | 107 | 0.003 | 58 | 0.005 | 32 | 0.007 |
| mo | 3 | 0.049 | 19 | 0.008 | 17 | 0.019 | 49 | 0.003 | 90 | 0.001 | 118 | 0.001 |
| mt | 123 | 0.003 | 18 | 0.022 | 11 | 0.019 | 16 | 0.017 | - | - | - | - |
| ne | - | - | - | - | 275 | 0.0 | 19 | 0.01 | 0 | 0.057 | 3 | 0.037 |
| nv | 20 | 0.01 | 33 | 0.006 | 35 | 0.006 | 101 | 0.002 | 2 | 0.072 | 6 | 0.044 |
| nh | 21 | 0.012 | 57 | 0.011 | 2 | 0.089 | 42 | 0.004 | 44 | 0.005 | 27 | 0.009 |
| nj | 66 | 0.004 | 352 | 0.0 | 363 | 0.0 | 14 | 0.01 | 397 | 0.0 | 340 | 0.0 |
| nm | 225 | 0.0 | 60 | 0.006 | 34 | 0.009 | 91 | 0.002 | 69 | 0.005 | 101 | 0.004 |
| ny | 167 | 0.002 | 20 | 0.012 | 20 | 0.014 | 36 | 0.004 | 47 | 0.007 | 49 | 0.006 |
| nc | 8 | 0.017 | 18 | 0.026 | 3 | 0.058 | 30 | 0.007 | 37 | 0.005 | 43 | 0.003 |
| nd | 121 | 0.002 | 13 | 0.016 | 14 | 0.02 | 12 | 0.014 | - | - | - | - |
| oh | 207 | 0.001 | 18 | 0.01 | 5 | 0.033 | 10 | 0.026 | 29 | 0.019 | 21 | 0.019 |
| ok | 116 | 0.0 | 82 | 0.0 | 21 | 0.001 | 66 | 0.46 | 134 | 0.0 | 174 | 0.0 |
| or | 32 | 0.012 | 34 | 0.006 | 27 | 0.007 | 4 | 0.025 | 40 | 0.007 | 2 | 0.029 |
| pa | 333 | 0.001 | 18 | 0.014 | 20 | 0.013 | 68 | 0.004 | 16 | 0.01 | 36 | 0.01 |
| ri | 126 | 0.002 | 16 | 0.022 | 24 | 0.008 | 42 | 0.006 | - | - | - | - |
| sc | 8 | 0.019 | 19 | 0.01 | 14 | 0.014 | 37 | 0.007 | - | - | - | - |
| sd | 227 | 0.0 | 46 | 0.006 | 27 | 0.0 | 7 | 0.042 | - | - | - | - |
| tn | 62 | 0.004 | 7 | 0.016 | 2 | 0.023 | 17 | 0.012 | 96 | 0.003 | 68 | 0.006 |
| tx | 147 | 0.003 | 212 | 0.001 | 61 | 0.003 | 22 | 0.014 | 112 | 0.001 | 108 | 0.002 |
| ut | - | - | - | - | - | - | 1 | 0.032 | - | - | 19 | 0.003 |
| vt | 46 | 0.004 | 53 | 0.003 | 44 | 0.004 | 2 | 0.084 | 20 | 0.019 | 50 | 0.097 |
| va | 4 | 0.048 | 59 | 0.003 | 139 | 0.001 | 101 | 0.002 | 136 | 0.001 | 222 | 0.001 |
| wa | 16 | 0.014 | 20 | 0.011 | 24 | 0.011 | 73 | 0.003 | 51 | 0.004 | 111 | 0.002 |
| wv | 35 | 0.004 | 7 | 0.018 | 6 | 0.042 | 34 | 0.003 | 2 | 0.099 | 7 | 0.021 |
| wi | 232 | 0.002 | 9 | 0.024 | 1 | 0.072 | 47 | 0.007 | 5 | 0.022 | 1 | 0.103 |
| wy | 162 | 0.0 | 49 | 0.004 | 48 | 0.004 | 85 | 0.002 | - | - | - | - |
| us | 20 | 0.016 | 6 | 0.029 | 6 | 0.025 | 10 | 0.01 | 16 | 0.02 | 16 | 0.013 |

Table 8: Feature ranks and weight for bipartisan, sponsor in minority, sponsor in majority, number of sponsors, average primary sponsor effectiveness and best primary sponsor effectiveness features in the gradient boosted model with just sponsor features across all states. Dash indicates feature was not ranked within the top 400.

| State | Top Phrases | Bottom Phrases |
|---|---|---|
| New Jersey (upper) | for farmland preserv, preserv trust, green acr fund, acquisit and, mmvv million from, vehicl from, budget for, fund for state, in feder fund, unemploy, for state acquisit, infrastructur trust | retir benefit for, of educ for, school board member, clarifi law, contract and, tax reimburs program, appropri mmvv for, to develop and, tax rate, credit under corpor, certain vehicl |
| New Jersey (lower) | environment infrastructur, to dissemin, farm to, dmva to, concern certain, and dhs, unsolicit, atm, contract law, link to, manufactur rebat, limit liabil | polit, import, all school, for water, facil to be, respons for, grant program for, relat crime, state administ, from tax, chair, to all |
| Maryland (upper) | financ the construct, festiv licens, issu the licens, grante provid and, to effect, advisori commiss, an evalu of, that provis of, financ statement, board licens, to borrow, defer | not to, phase, be use as, use as, facil locat in, to own, law petit, or expenditur, trust establish, expend match, and expend match |
| Maryland (lower) | charl counti alcohol, improv or, to financ the, termin provis relat, counti sale, sanction, alter, counti special tax, montgomeri counti alcohol, report requir repeal, length, licens mc | grant to the, creation state debt, educ fund, state debt baltimor, establish the amount, elimin, disclos to, propos amend, incom tax rate, purpos relat, crimin gang, deced die after |
| California (upper) | ab, revolv fund, household, intent that, these provis until, restitut, counsel, employe, onli if ab, properti, if ab, would incorpor addit | legisl, cost of, veterinari, enact legisl, to the, law, regul econom, governor, incom tax deduct, hour, motor vehicl recreat, decis |
| California (lower) | add articl, to amend repeal, bill would incorpor, to add and, budget act of, urgenc statut, make nonsubstant, and make, as bill provid, relat the budget, and of the, ab | would make nonsubstant, enact legisl, make technic nonsubstant, would make technic, unspecifi, code to add, baccalaur degre, salari, fraud prevent, flexibl, of the state, would |
| Florida (upper) | ogsr, abrog provis relat, grant trust fund, govern act, person inform, to supplement, employ contribut to, legisl audit committe, jac, maintain by the, insur regul, financi inform | senat relat to, senat relat, to, ssb, elder, school, municip that, and legislatur by, that law enforc, provid minimum, admiss to, local law enforc |
| Florida (lower) | etc, certain propos, re creat, repeal under, to qualifi, boundari, program revis requir, environment permit, counti hospit district, alcohol beverag licens, except under, ranch | hous relat, day, renew energi, provid for alloc, make recommend, for employ of, of damag, from particip, week, catastroph, dhsmv to develop, employ from |
| Delaware (upper) | uniform, would increas the, amend chapter volum, person convict, relat the delawar, dealer, child support, bureau, violenc, associ, charter chang, for fiscal year | rent, state languag, the content, act regul, certain licens, give local, assembl from, delawar code establish, for citizen, reimburs, propos constitut amend, salari |
| Delaware (lower) | of the th, tax refund, thi act also, amend of the, this section of, the titl, the act to, of member of, electron transmiss, for in the, and the date, parent guardian | predatori, hour per, relat state employe, unfair practic, communic, open meet, equal the, to the construct, the construct, medicaid, state agenc, relat to prevail |

Table 9: Top and bottom ranked phrases for New Jersey, Maryland, California, and Florida.